%% file: main.tex
\documentclass[10pt,twocolumn,letterpaper]{article}

\pdfoutput=1

\usepackage[pagenumbers]{cvpr}
\usepackage{graphicx}
\usepackage{amsmath}
\usepackage{amssymb}
\usepackage{booktabs}
\usepackage{cite}
\usepackage{multirow}
\usepackage{colortbl}
\usepackage[linesnumbered,lined,vlined,ruled,commentsnumbered,noend]{algorithm2e}
\usepackage[pagebackref=true, breaklinks=true, colorlinks,
 citecolor=citecolor, linkcolor=linkcolor, bookmarks=false]{hyperref}
\usepackage[capitalize]{cleveref}
\usepackage{xcolor}
\crefname{section}{Sec.}{Secs.}
\Crefname{section}{Section}{Sections}
\Crefname{table}{Table}{Tables}
\crefname{table}{Tab.}{Tabs.}
\Crefname{equation}{Equation}{Equations}
\crefname{equation}{eq.}{eqs.}

\definecolor{citecolor}{HTML}{0071BC}
\definecolor{linkcolor}{HTML}{ED1C24}

\input{math_commands}

\makeatletter
\renewcommand{\paragraph}{%
  \@startsection{paragraph}{4}%
  {\z@}{0.25em}{-1em}%
  {\normalfont\normalsize\bfseries}%
}
\makeatother

\definecolor{mygray}{gray}{0.6}
\definecolor{mygray-bg}{gray}{0.9}

\begin{document}
\title{Online Hyperparameter Optimization for Class-Incremental Learning}
\author{Yaoyao Liu$^{1}$ 
\quad Yingying Li$^{2}$ 
\quad Bernt Schiele$^{1}$
\quad Qianru Sun$^{3}$\\
\\
\small $^{1}$Max Planck Institute for Informatics, Saarland Informatics Campus\\
\small $^{2}$Computing and Mathematical Sciences, California Institute of Technology\\
\small $^{2}$School of Computing and Information Systems, Singapore Management University\\
\small {\texttt{\{yaoyao.liu, schiele\}@mpi-inf.mpg.de}}  \quad  \texttt{yingli2@caltech.edu}  \quad \texttt{qianrusun@smu.edu.sg}
}

\newcommand{\myparagraph}[1]{\noindent\textbf{#1}}
\definecolor{mygray}{gray}{0.6}
\definecolor{mygray-bg}{gray}{0.9}

\newcommand{\citefirst}[1]{%
  \AtNextCite{\defcounter{maxnames}{1}\defcounter{minnames}{1}}%
  \citet{#1}%
}

\maketitle

\input{sections/0_abstract}
\input{sections/1_intro}
\input{sections/2_related_work}

\input{sections/3_preliminaries}
\input{sections/4_method}
\input{sections/5_experiments}
\input{sections/6_conclusion}

{\small\bibliographystyle{ieee_fullname}\bibliography{aaai23}}

\clearpage
\input{sections/7_supplementary}

\end{document}

%% file: math_commands.tex
\usepackage{amsmath,amsfonts,bm}

\def\eqref#1{equation~\ref{#1}}

\def\1{\bm{1}}

\DeclareMathAlphabet{\mathsfit}{\encodingdefault}{\sfdefault}{m}{sl}
\SetMathAlphabet{\mathsfit}{bold}{\encodingdefault}{\sfdefault}{bx}{n}



%% file: sections/0_abstract.tex
\begin{abstract}

Class-incremental learning (CIL) aims to train a classification model while the number of classes increases phase-by-phase. An inherent challenge of CIL is the stability-plasticity tradeoff, i.e., CIL models should keep stable to retain old knowledge and keep plastic to absorb new knowledge. However, none of the existing CIL models can achieve the optimal tradeoff in different data-receiving settings---where typically the training-from-half (TFH) setting needs more stability, but the training-from-scratch (TFS) needs more plasticity.
To this end, we design an online learning method that can adaptively optimize the tradeoff without knowing the setting as a priori.
Specifically, we first introduce the key hyperparameters that influence the tradeoff, e.g., knowledge distillation~(KD) loss weights, learning rates, and classifier types.
Then, we formulate the hyperparameter optimization process as an online Markov Decision Process (MDP) problem and propose a specific algorithm to solve it. 
We apply local estimated rewards and a classic bandit algorithm Exp3~\cite{auer2002nonstochastic} to address the issues when applying online MDP methods to the CIL protocol.
Our method consistently improves top-performing CIL methods in both TFH and TFS settings, e.g., boosting the average accuracy of TFH and TFS by 2.2 percentage points on ImageNet-Full, compared to the state-of-the-art~\cite{Liu2021RMM}.\footnote{Code: \href{https://class-il.mpi-inf.mpg.de/online/}{https://class-il.mpi-inf.mpg.de/online/}}
\end{abstract}

%% file: sections/1_intro.tex
\section{Introduction}
\label{sec_intro}

\input{figures/1_teaser_new}

Real-world problems are ever-changing, with new concepts and new data being continuously observed. Ideal AI systems should have the ability to learn new concepts from the new data, also known as \textit{plasticity}, while maintaining the ability to recognize old concepts, also known as \textit{stability}. However, there is a fundamental \textit{tradeoff} between plasticity and stability: too much plasticity may result in the \textit{catastrophic forgetting} of old concepts, while too much stability restricts the ability to adapt to new concepts~\cite{mccloskey1989catastrophic, McRae1993Catastrophic, Ratcliff1990catastrophic}. 
To encourage related research, Rebuffi et al.~\cite{rebuffi2017icarl} defined the class-incremental learning (CIL) protocol, where the training samples of different classes come to the model phase-by-phase and most of the past data are removed from the memory.

Recently, many methods have been proposed to balance the stability-plasticity tradeoff for \textit{different data-receiving settings} of CIL. For example, the strong feature knowledge distillation (KD) loss function is usually adopted when a large amount of data is available beforehand (e.g., the \emph{training-from-half} (TFH) setting) since it encourages stability
~\cite{hou2019lucir,Liu2020AANets,Liu2021RMM}; while the weak logit KD loss function is popular when data and classes are received evenly in each phase (e.g., the \emph{{training-from-scratch}} (TFS) setting) since it provides more plasticity~\cite{rebuffi2017icarl,belouadah2019il2m,Li18LWF}. 

Most CIL algorithms pre-fix the tradeoff balancing methods, usually according to which data-receiving setting will be used in the experiments.  However, in real-world scenarios, it is difficult to anticipate how data will be received in the future. Hence, a pre-fixed method is no longer proper for balancing the stability and plasticity of the actual data stream, thus generating worse performance. This can also be validated in Figure~\ref{intro_figure}. Notice that the methods with weak KD, e.g., iCaRL~\cite{rebuffi2017icarl} and LwF~\cite{Li18LWF}, provide worse performance in TFH than in TFS since weak KD provides too much plasticity for TFH, while the methods with strong KD, e.g., LUCIR~\cite{hou2019lucir}, AANets~\cite{Liu2020AANets}, and RMM~\cite{Liu2021RMM}, perform worse in TFS than in TFH due to too much stability.

Therefore, a natural question is: how to design an \textit{adaptive trade-off balancing method} to achieve good performance  \textit{without knowing} how data will be received \textit{beforehand}?
To tackle this, we propose an online-learning-inspired method to adaptively adjust  key hyperparameters that affect the trade-off balancing performance in CIL. 
In our method, we introduce hyperparameters to control the choice of KD loss functions, learning rates, and  classifier types, which are key algorithm choices that affect the tradeoff balancing performance.\footnote{The impact of KD losses has been discussed before. Learning rates naturally affect how fast the model learns new concepts.  We also adjust the classifier type because empirical results~\cite{rebuffi2017icarl,hou2019lucir} show that the nearest class mean (NCM) and fully-connected (FC) classifiers perform better under more plasticity and stability, respectively.} In this way, deciding this choice is transformed into a hyperparameter optimization (HO) problem. 

This HO problem cannot be directly solved because future data are not available. Thus, we borrow ideas from  online learning, which is a widely adopted approach to adaptively tune the decisions without knowing  the future data a priori while still achieving good performance in hindsight. 

However, in CIL, our decisions affect not only the next phase but also all the future phases, which is different from the standard online learning setting~\cite{anderson2008theory}. 
 To capture the dependence across phases, we formulate the HO problem in CIL as an online MDP, which is a generalized version of online learning. Further, we propose a new algorithm based on~\cite{even2009online} to solve this online MDP problem. Our algorithm differs from the standard online MDP algorithm in \cite{even2009online} in two aspects:
\begin{itemize}
    \item In CIL, we cannot directly observe the reward (i.e., validation accuracy) because the validation data is not accessible during training. To address this issue, we estimate the reward by rebuilding  local training and validation sets during policy learning in each phase, and computing the estimated reward on the local validation sets.
    
    \item In CIL, we only have access to the model generated by the selected hyperparameters instead of the models generated by other hyperparameters. In other words, we only have bandit feedback instead of full feedback as assumed in \cite{even2009online}. To address this, we revise the algorithm in \cite{even2009online} by combining it with a classic bandit algorithm, Exp3 \cite{auer2002nonstochastic}.

\end{itemize}

Empirically, we find our method performs well consistently. We conduct extensive CIL experiments by plugging our method into three top-performing methods (LUCIR~\cite{hou2019lucir}, AANets~\cite{Liu2020AANets}, and RMM~\cite{Liu2021RMM}) and testing them on three benchmarks (i.e., CIFAR-100, ImageNet-Subset, and ImageNet-Full). Our results show the consistent improvements of the proposed method, e.g., boosting the average accuracy of TFH and TFS by $2.2$ percentage points on ImageNet-Full, compared to the state-of-the-art~\cite{Liu2021RMM}.

Lastly, it is worth mentioning that our method can also be applied to optimize other key hyperparameters in CIL, e.g., memory allocation~\cite{Liu2021RMM}.

\textbf{Summary of our contributions.} Our contributions are three-fold: 1)~an online MDP formulation that allows online updates of hyperparameters that affect the balance of the stability and plasticity in CIL; 2)~an Exp3-based online MDP algorithm to generate adaptive hyperparameters using bandit and estimated feedback; 3)~extensive comparisons and visualizations for our method in three CIL benchmarks, taking top-performing methods as baselines.

%% file: figures/1_teaser_new.tex
\begin{figure}
\centering
\includegraphics[width=2.8in]{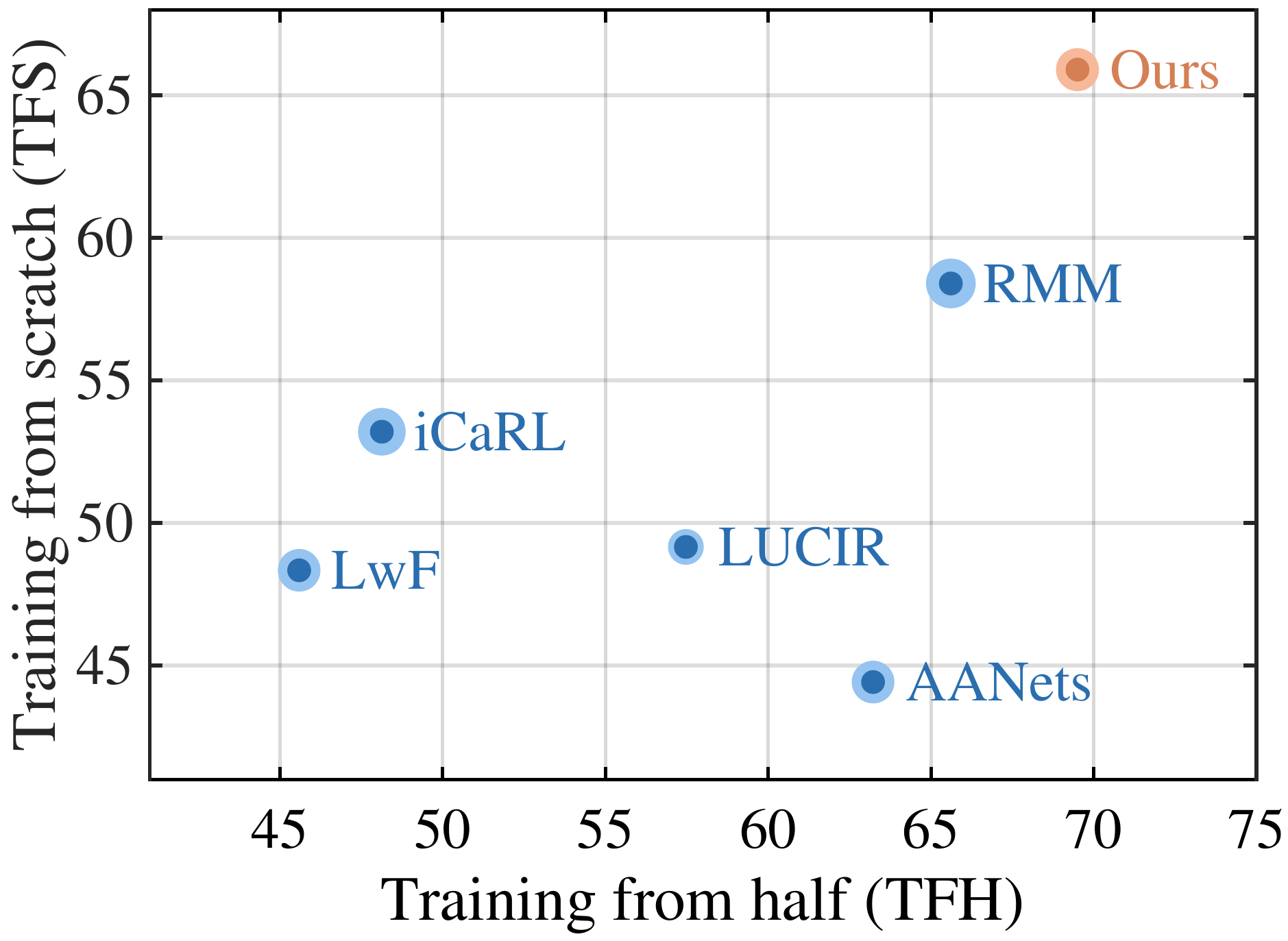}
\caption{
Average accuracy (\%) on CIFAR-100 25-phase, using two data-receiving settings: 1) \emph{training-from-half}~(TFH): a large amount of data is available beforehand to pre-train the encoder; 2) \emph{training-from-scratch} (TFS): 
classes come evenly in each phase.
\textcolor[rgb]{0.165, 0.431, 0.686}{\textbf{Dark blue}} and \textcolor[rgb]{0.835,0.498,0.33}{\textbf{orange}} indicate the baselines and our method, respectively. Light-color circles are confidence intervals. Notice that methods with strong KD losses, e.g., LUCIR~\cite{hou2019lucir}, AANets~\cite{Liu2020AANets}, and RMM~\cite{Liu2021RMM}, tend to provide worse performance in TFS than TFH, while methods with weak KD losses, e.g., iCaRL~\cite{rebuffi2017icarl} and LwF~\cite{Li18LWF}, tend to provide worse performance in TFH than TFS. Our method uses an online learning algorithm to produce the key hyperparameters, e.g., the weights that control which KD losses are used. Thus, our method achieves the highest performance in both TFS and TFH.
}
\label{intro_figure}
\end{figure}

%% file: sections/2_related_work.tex
\section{Related Work}
\label{sec2_related_work}

\input{figures/2_global_flow}

\myparagraph{Class-incremental learning (CIL).}  
There are three main lines of work to address the stability-plasticity trade-off in CIL. {\emph{{Distillation-based}}} methods~\cite{Joseph2022Energy,Li18LWF,douillard2020podnet,Pourkeshavarzi2022Looking,TrikiABT17,yu2020CVPRsemantic,hu2021cil} introduce different knowledge distillation (KD) losses to consolidate previous knowledge when training the model on new data. The key idea is to enforce model prediction logits~\cite{Li18LWF,rebuffi2017icarl}, feature maps~\cite{douillard2020podnet,hou2019lucir}, or topologies in the feature space~\cite{Tao2020topology} to be close to those of the pre-phase model. {\emph{{Memory-based}}} methods~\cite{rebuffi2017icarl,shin2017continual,liu2020mnemonics,prabhu12356gdumb,belouadah2019il2m} preserve a small number of old class data (called exemplars) and train the model on them together with new class data. {\emph{{Network-architecture-based}}} methods~\cite{rusu2016progressive,xu2018reinforced,abati2020conditional,yan2021DER,wang2022foster}  design incremental network architectures by expanding the network capacity for new data or freezing partial network parameters to keep the knowledge of old classes. However, all the above methods are either too stable or too plastic to perform well in both TFS and TFH settings. Our method learns an online policy to generate hyperparameters that balance stability and plasticity. Therefore, our method performs well in both TFS and TFH settings.

\myparagraph{Reinforcement learning (RL)} 
aims to learn a policy in an environment, which is typically formulated as an MDP. Some CIL papers also deploy RL algorithms in their frameworks. Xu et al.~\cite{xu2018reinforced} used RL to expand its backbone network when a new task arrives adaptively. Liu et al.~\cite{Liu2021RMM} used RL to learn a policy to adjust the memory allocation between old and new class data dynamically along with the learning phases. Our method focuses on learning a policy to produce key hyperparameters. 
Besides, the existing methods need to solve the complete MDP, which is time-consuming. Here, we formulate the CIL task as an online MDP. Thus, our method is more time-efficient. 

\myparagraph{Online learning} observes a stream of samples and makes a prediction for each element in the stream. There are mainly two settings in online learning: {full feedback} and {bandit feedback}. {\emph{{Full feedback}}} means that the full reward function is given at each stage. It can be solved by Best-Expert algorithms~\cite{even2005experts}. {\emph{{Bandit feedback}}} means that only the reward of the implemented decision is revealed. If the  rewards  are independently drawn from a fixed and unknown distribution, we may use e.g., Thompson sampling~\cite{agrawal2012analysis} and UCB~\cite{auer2010ucb} to solve it. If the rewards are generated in a non-stochastic version, we can solve it by, e.g., Exp3~\cite{auer2002nonstochastic}. {\emph{{Online MDP}}} is an extension of online learning. Many studies~\cite{even2009online,li2019online3,li2019online2,li2021online} aim to solve it by converting it to online learning. In our case, we formulate the CIL as an online MDP and convert it into a classic online learning problem. The rewards in our MDP are non-stochastic because the training and validation data change in each phase. Therefore, we design our algorithm based on Exp3~\cite{auer2002nonstochastic}.

\myparagraph{Hyperparameter optimization (HO).}  There are mainly two popular lines of HO methods: gradient-based and meta-learning-based. 
{Gradient-based HO methods~\cite{baydin2018automatic} make it possible to tune the entire weight vectors associated with a neural network layer as hyperparameters. Meta-learning-based HO methods~\cite{franceschi2018bilevel} use a bilevel program to optimize the hyperparameters. However, all these methods only consider time-invariant environments. Our online method learns hyperparameters that adapt to the time-varying environments in CIL.
}

%% file: figures/2_global_flow.tex
\begin{figure*}[t]

\includegraphics[width=\textwidth]{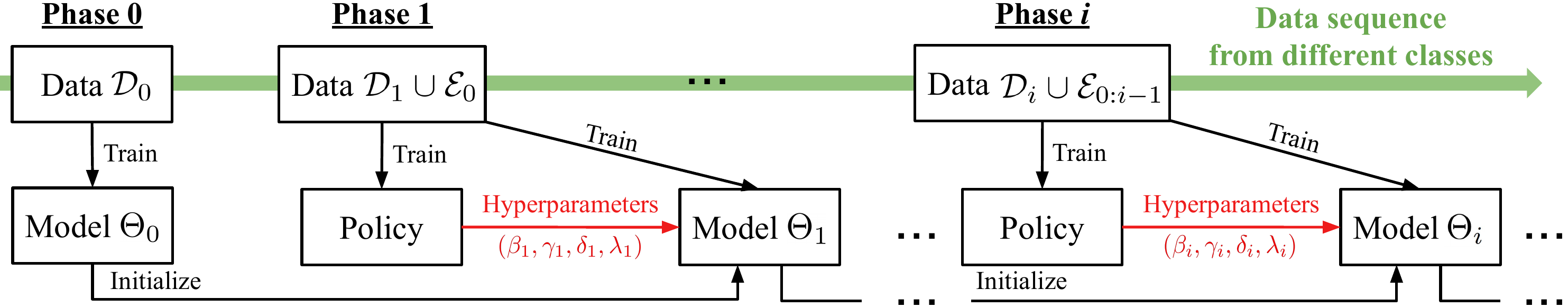}
\caption{The computing flow of our method. We formulate the CIL task as an online MDP: each phase in CIL is a stage in the MDP, and the CIL models are the states. We train the policy to produce actions, which contain the hyperparameters we use in the CIL training. We illustrate the training process of each phase in Figure~\ref{framework_figure}.}
  \label{global_figure}
\end{figure*}

%% file: sections/3_preliminaries.tex
\section{Preliminaries}
\label{sec3_preliminaries}

\myparagraph{Class-incremental learning (CIL).} The general CIL pipeline is as follows. There are multiple phases during which the number of classes gradually increases to the maximum~\cite{douillard2020podnet,hou2019lucir,hu2021cil,liu2020mnemonics}. In the {$0$-th} phase, we observe data $\mathcal{D}_{0}$, and use it to learn an initial model $\Theta_0$. After this phase, we can only store a small subset of $\mathcal{D}_{0}$ (i.e., exemplars denoted as $\mathcal{E}_{0}$) in memory used as replay samples in later phases. In the $i$-th phase ({$i$}{$\geq$}{$1$}), we get new class data $\mathcal{D}_{i}$ and load exemplars $\mathcal E_{0:i-1}$=$\mathcal E_0\cup \dots \cup \mathcal E_{i-1}$ from the memory. Then, we initialize $\Theta_i$ with $\Theta_{i-1}$, and train it using $\mathcal E_{0:i-1}\cup\mathcal{D}_{i}$. We evaluate the model $\Theta_{i}$ on a test set $\mathcal{Q}_{0:i}$ for all classes observed so far. After that, we select exemplars $\mathcal E_{0:i}$ from $\mathcal E_{0:i-1}\cup\mathcal{D}_{i}$ and save them in the memory.

Existing work mainly focus on two data-receiving settings: \emph{training-from-half} (TFH)~\cite{hou2019lucir} and {\emph{training-from-scratch}} (TFS)~\cite{rebuffi2017icarl} settings.
In TFH, we are given half of all classes in the $0$-th phase, and observe the remaining classes evenly in the subsequent $N$ phases. In TFS, we are given the same number of classes in all $N$ phases.

%% file: sections/4_method.tex
\section{Methodology}
\label{sec4_method}

As illustrated in Figures~\ref{global_figure} and~\ref{framework_figure}, we formulate  CIL as an online MDP and learn a policy to produce the hyperparameters in each phase. In this section, we introduce the
online MDP formulation, show optimizable hyperparameters, and provide an online learning algorithm to train the policy. 
Algorithm~\ref{algo_Exp3} summarizes the overall training steps of the proposed method. Algorithm~\ref{algo_CIL} presents the training and evaluation process for a specific action.

\input{figures/3_framework}

\subsection{An Online MDP Formulation for CIL}
\label{sec4_1_mdp}

Optimizing hyperparameters in CIL should be online inherently: training and validation data changes in each phase, so the hyperparameters should be adjusted accordingly. Thus, it is intuitive to formulate the CIL as an online MDP. In the following, we provide detailed formulations.

\myparagraph{Stages.} Each phase in the CIL task can be viewed as a stage in the online MDP. 

\myparagraph{States.} The state should define the current situation of the intelligent agent. In CIL, we use the  model $\Theta_i$ as the state of the $i$-th phase/stage. We use $\mathbb S$ to denote the state space.

\myparagraph{Actions.} We use a 
vector consisting of the hyperparameters in the $i$-th phase as the action $\mathbf{a}_i$. When we take the action $\mathbf{a}_i$, we  deploy the corresponding hyperparameters. We denote the action space as $\mathbb A$. Please refer to the next subsection, \emph{{Optimizable Hyperparameters}}, for more details.

\myparagraph{Policy} {$\mathbf p$}{$=$}$\{p({\mathbf{a}|\Theta_i})\}_{\mathbf{a}\in\mathbb A}$ is a probability distribution over the action space $\mathbb A$, given the current state $\Theta_i$.

\myparagraph{Environments.} We define the training and validation data in each phase as the environment. In the $i$-th phase, the environment is {$\mathcal{H}_i$}{$=$}$(\mathcal E_{0:i-1}\cup\mathcal{D}_{i}, \mathcal Q_{0:i})$, where $\mathcal E_{0:i-1}\cup\mathcal{D}_{i}$ is the training data and $\mathcal Q_{0:i}$ is the corresponding validation data. The environment is time-varying because we are given different training and validation data in each phase. 

\myparagraph{Rewards.}  CIL aims to train a model that is efficient in recognizing all classes seen so far. Therefore, we use the validation accuracy as the reward in each phase. Our objective is to maximize a cumulative reward, i.e., $R=\sum_{i=1}^{N}r_{\mathcal{H}_i}(\Theta_i,\mathbf{a}_i)$, where $r_{\mathcal{H}_i}(\Theta_i,\mathbf{a}_i)$ denotes the $i$-th phase reward, i.e., the validation accuracy of $\Theta_i$. The reward function $r_{\mathcal{H}_i}$ changes with ${\mathcal{H}_i}$, so it is time-varying. 

\subsection{Optimizable Hyperparameters}
\label{sec4_2_hyper}

In this part, we introduce the optimizable hyperparameters, and how to define the actions and action space based on the hyperparameters.
We consider three kinds of hyperparameters that significantly affect stability and plasticity: 1) KD loss weights, 2) learning rates, and 3) classified types. 

\myparagraph{1) KD loss weights.} We first introduce two KD losses (i.e., logit and feature KD losses) and then show how to use the KD loss weights to balance them.

\myparagraph{\emph{{Logit KD loss}}} is proposed in~\cite{Hinton15KnowledgeDistillation} and widely applied in CIL methods~\cite{Li18LWF,rebuffi2017icarl,Liu2020AANets}. Its motivation  is to make the current model $\Theta_i$ mimic the prediction logits of the old model $\Theta_{i-1}$:
\begin{equation}
\label{eq_logit_distillation}   
\mathcal{L}_{\text{logi}}=-\sum_{k=1}^{K}\eta_k(\mu(x; \Theta_{i-1}))\log \eta_k(\mu(x; \Theta_{i})),
\end{equation}
where $\mu(x; \Theta)$ is a function that maps the input mini-batch $(x,y)$ to the prediction logits using the model $\Theta$. $\eta_k(v)=v_k^{{1}/{\tau}}/\sum_{j}v_j^{{1}/{\tau}}$ is a re-scaling function for the $k$-th class prediction logit and $\tau$ is a scalar set to be greater than $1$.

\myparagraph{\emph{{Feature KD loss}}}~\cite{hou2019lucir,douillard2020podnet} aims to enforce a stronger constraint on the previous knowledge by minimizing the cosine similarity between the features from the current model $\Theta_i$ and the old model $\Theta_{i-1}$. It can be computed as follows,
\begin{equation}
\label{eq_feature_distillation}   
\mathcal{L}_{\text{feat}}=1-S_c( f(x; \Theta_{i}), f(x; \Theta_{i-1})),
\end{equation}
where $f(x; \Theta)$ denotes a function that maps the input image $x$ to the features using the model $\Theta$. $S_c(v_1, v_2)$ denotes the cosine similarity between $v_1$ and $v_2$.

\myparagraph{\emph{{The overall loss}}} in the $i$-th phase is a weighted sum of the classification and different KD losses: 
\begin{equation}
\label{eq_overall_loss}  
\mathcal{L}_{\text{overall}} = \mathcal{L}_{\text{CE}} + \beta_i \mathcal{L}_{\text{logi}} + \gamma_i \mathcal{L}_{\text{feat}},
\end{equation}
where $\beta_i$ and $\gamma_i$ are the weights of the logit and feature KD losses, respectively. $\mathcal{L}_{\text{CE}}$ is the standard cross-entropy classification loss. Existing methods~\cite{Li18LWF,rebuffi2017icarl,hou2019lucir,Liu2020AANets,Liu2021RMM} can be viewed as using fixed heuristic KD loss weights, e.g., {$\beta_i$}$\equiv${$1$} and {$\gamma_i$}$\equiv${$0$} in iCaRL~\cite{Li18LWF,rebuffi2017icarl}. Instead, our method optimizes $\beta_i$ and $\gamma_i$ online. Thus, we can balance the model’s stability and plasticity by adjusting $\beta_i$ and $\gamma_i$. We apply different weights for the logit and feature KD losses so that we can achieve fine-grained control over the intensity of knowledge distillation.

\myparagraph{2) Learning rate} is another important hyperparameter that affects the model’s stability and plasticity. We empirically find that a lower learning rate makes the CIL model more stable, while a higher learning rate makes the CIL model more plastic. If we use $\lambda_i$ to denote the learnable learning rate in the $i$-th phase, we update the CIL model as follows,
\begin{equation}
    \Theta_i \gets \Theta_i - \lambda_i \nabla_{\Theta_i}\mathcal{L}_{\text{overall}}.
\label{eq_gd_lr} 
\end{equation}
Another hyperparameter, the number of training epochs, has similar properties to the learning rate. We choose to optimize the learning rate and fix the number of epochs because it empirically works better with our online learning algorithm.

\myparagraph{3) Classifier type.} Motivated by the empirical analysis, we consider two classifier types in our study: {\emph{nearest class mean}} (NCM)~\cite{rebuffi2017icarl,snell2017prototypical} and {\emph{fully-connected}} (FC)~\cite{hou2019lucir,liu2020mnemonics} classifiers. For the NCM classifier, we first compute the mean feature for each class using the new data and old exemplars. Then we perform a nearest neighbor search using the Euclidean distance on the $L_2$ normalized mean features to get the final predictions. It is observed empirically that the NCM classifier tends to work better on the models with high plasticity, while the FC classifier performs better on the models with high stability~\cite{rebuffi2017icarl,hou2019lucir}. Thus, we propose to use a hyperparameter, classifier type indicator $\delta_i$, to control the final predictions during the evaluation: 
\begin{equation}
\label{eq_pred}  
\mu(x; \Theta_i) =  \mu_{\text{ncm}}(x; \Theta_i) [\delta_i=1] +  \mu_{\text{fc}}(x; \Theta_i) [\delta_i=0], 
\end{equation}
where $\delta_i\in\{0,1\}$, $\mu_{\text{ncm}}$ and $\mu_{\text{fc}}$ are the predictions on the input image $x$ using the NCM and FC classifiers, respectively.

\myparagraph{Summary: actions and action space.} 
In summary, we define the action as {$\mathbf{a}_i$}{$=$}$(\beta_i, \gamma_i, \lambda_i, \delta_i, )$, which consists of the following hyperparameters: KD loss weights $\beta_i$ and $\gamma_i$, learning rate $\lambda_i$, and classifier type indicator $\delta_i$.
For the hyperparameters that may vary in a continuous range, we \emph{discretize} them to define a \emph{finite} action space.\footnote{Though discretization suffers the curse of dimensionality, our experiments show that with a coarse grid, we already have significant improvements over pre-fixed hyperparameters.} In the next subsection, we show how to learn the policy in each phase.

\subsection{Policy Learning}
\label{sec4_3_policy}

A common approach to solving an online MDP is to approximate it as an online learning problem and solve it using online learning algorithms~\cite{even2005experts,agrawal2012analysis,auer2002nonstochastic}. We also take this approach, and our approximation follows~\cite{even2009online}, which achieves the optimal regret. In their paper, Even-Dar et al.~\cite{even2009online} relax the Markovian assumption of the MDP by decoupling the cumulative reward function and letting it be time-dependent so that they can solve the online MDP by standard online learning algorithms. {Such
a decoupling requires the following assumptions. 1) Fast mixing: in CIL, the hyperparameters in an early phase do not have much impact on the test accuracy of the classes observed in the current phase. 2) The algorithm changes the hyperparameters slowly (this can be observed in \textbf{\emph{Experiments}} \& Figure~\ref{figure_values}). Thus, these assumptions fit our CIL problem.}

However, we cannot directly apply the algorithms proposed in \cite{even2009online} to our problem. It is because their paper assumes \textit{full feedback}, i.e., we can observe the rewards of all actions in each phase. Therefore, its online learning problem could be solved by Best Expert algorithms~\cite{even2005experts}.
In CIL, we cannot observe any reward (i.e., validation accuracy) because the validation data is not accessible during training. To address this issue, we rebuild the local training and validation sets in each phase. In this way, our problem has \textit{bandit feedback}: we can compute the reward of the implemented action. Therefore, we can solve our online learning problem based on Exp3~\cite{auer2002nonstochastic}, a famous bandit algorithm.

In the following, we show how to rebuild the local training and validation sets, compute the decoupled cumulative reward, and learn the policy with Exp3.

\myparagraph{Rebuilding local datasets.} In the $i$-th phase, we need to access the validation set $\mathcal Q_{0:i}$ to compute the reward (i.e., the validation accuracy). However, we are not allowed to use $\mathcal Q_{0:i}$ during training because it violates the CIL benchmark protocol. Therefore, we replace $\mathcal Q_{0:i}$ with a class-balanced subset $\mathcal B_{0:i}$ sampled from the training data $\mathcal E_{0:i-1}\cup\mathcal{D}_{i}$. 
$\mathcal B_{0:i}$ contains the same number of samples for both the old and new classes.
In this way, we can rebuild the local training and validation sets, and obtain the local environment $h_i=((\mathcal E_{0:i-1}\cup\mathcal{D}_{i})\setminus\mathcal B_{0:i}, \mathcal B_{0:i})$.

\myparagraph{Decoupled cumulative reward.} 
We create the decoupled cumulative reward function $\hat{R}$ based on the original cumulative reward function ${R}=\sum_{j=1}^{N}r_{{\mathcal{H}}_j}(\Theta_j,\mathbf{a}_j)$. In the $i$-th phase, we compute  $\hat{R}$ as follows,
\begin{equation}
\label{eq_reward_1}
\hat{R}(\mathbf{a}_i, h_i)=\underbrace{\sum_{j=1}^{i-1}r_{{\mathcal{H}}_j}(\Theta_j,\mathbf{a}_j)}_{\text{Part I}}+\underbrace{\sum_{j=i}^{i+n}r_{h_i}(\Theta_j,\mathbf{a}_i)}_{\text{Part II}},
\end{equation}
where Part~I is the historical rewards from the $1$-st phase to the $(i$-$1)$-th phase. It is a constant and doesn't influence policy optimization. Part II is the long-term reward of a time-invariant local MDP based on the local environment $h_i$. We use Part~II as an estimation of the future rewards, following~\cite{even2009online}. Because we don't know the total number of phases $N$ during training, we assume there are $n$ phases in the future.  Furthermore, we fix the action $\mathbf{a}_i$ in Part~II to simplify the training process. Thus, $\hat{R}$ can be reviewed as a function of $\mathbf{a}_i$ and $h_i$.

\myparagraph{Training policy with Exp3.} 
Exp3~\cite{auer2002nonstochastic} introduces an auxiliary variable $\mathbf w=\{w({\mathbf{a}})\}_{\mathbf{a}\in\mathbb A}$. After updating $\mathbf w$, we can determine the policy {$\mathbf p$}{$=$}$\{p({\mathbf{a}|\Theta_i})\}_{\mathbf{a}\in\mathbb A}$ by 
\begin{equation}
\label{eq_exp3_p}
\mathbf p={\mathbf w}/{||\mathbf w||}.
\end{equation}
The updating rule of w is provided below. 

In the $1$-st phase, we initialize $\mathbf w$ as $\{1,\dots,1\}$. In each phase~$i$ ({$i$}{$\geq$}{$1$}), we update $\mathbf w$ for $T$ iterations. In the $t$-th iteration, we sample an action {$\mathbf{a}_t$}{$\sim$}{$\mathbf p$}, apply the action $\mathbf{a}_t$ to the CIL system,  and compute the decoupled cumulative reward $\hat{R}(\mathbf{a}_t, h_i)$ using Eq.~\ref{eq_reward_1}. After that, we update $w({\mathbf{a}}_t)$ in $\mathbf w$ as,
\begin{equation}
\label{eq_exp3_update}
w({\mathbf{a}}_t) \gets w({\mathbf{a}_t})\exp(\xi{\hat{R}(\mathbf{a}_t, h_i)}/{p(\mathbf{a}_t|\Theta_i)}),
\end{equation}
where $\xi$ is a constant, which can be regarded as the learning rate in Exp3. 

\input{pseudocode/1_Exp3}
\input{pseudocode/2_CIL}

%% file: figures/3_framework.tex
\begin{figure*}[t]

\includegraphics[width=\textwidth]{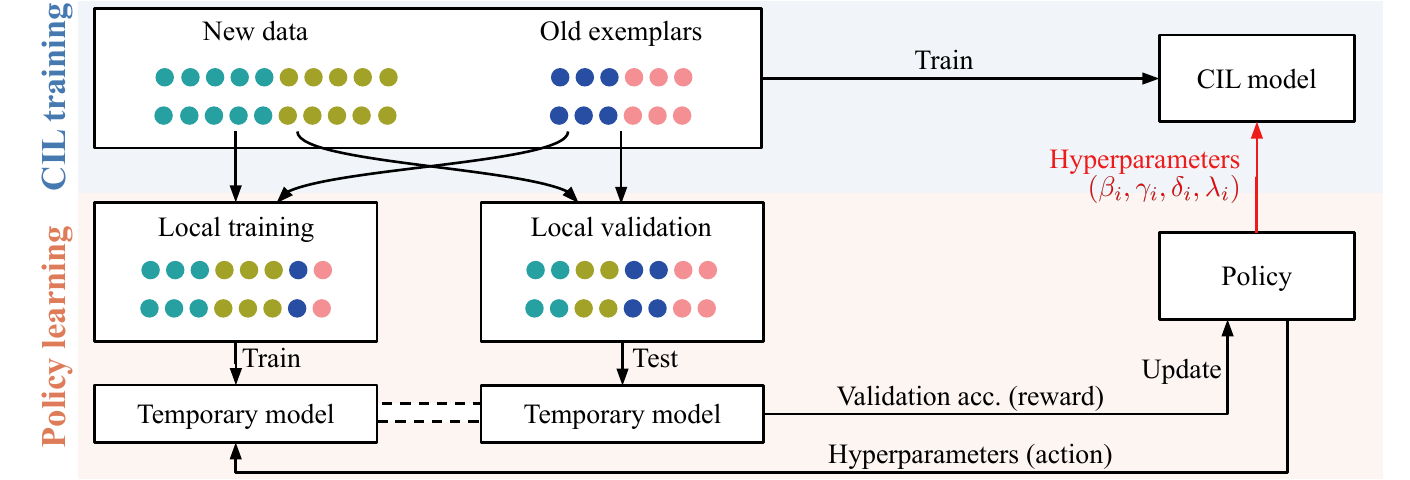}
\caption{The training process of our online learning method in the $i$-th phase. 
It includes {\emph{policy learning}} and {\emph{CIL training}}.
\textbf{(a)~Policy learning}. 1) We construct a class-balanced subset from all training data as the local validation set and use the remaining data as the local training set. 2) We initialize the temporary model with $\Theta_{i-1}$. 3) We sample an action using the current policy and deploy the hyperparameters on the temporary model according to the action. 4) We train it on the local training set for $M_1$ epochs, and evaluate it on the local test set. 5) We use the validation accuracy as the reward and update the policy. We update the policy for $T$ iterations by repeating Steps 2-5. \textbf{(b) CIL training}. We sample an action using the learned policy and deploy the hyperparameters on the CIL model. Then, we train the CIL model on all training data for $M_2$ epochs. 
We set $M_1=\lceil 0.1M_2 \rceil$ to speed up the policy learning.
}
  \label{framework_figure}
\end{figure*}

%% file: pseudocode/1_Exp3.tex
\begin{algorithm}[t]
 \DontPrintSemicolon
 \SetKwInOut{Input}{Input}
 \SetKwInOut{Output}{Output}
 \Input{Old model $\Theta_{i-1}$, training data $\mathcal E_{0:i-1}\cup\mathcal{D}_{i}$, validation data $\mathcal Q_{0:i}$, \\learnable parameters $\mathbf w$, numbers of epochs $M_1$ and $M_2$.
 }
 \Output{New model $\Theta_{i}$, new exemplars $\mathcal E_{0:i}$, learnable parameters $\mathbf w$.}
 {\color[rgb]{0.835,0.498,0.33} \tcp{Policy learning}}
 \If{\rm $i$=$1$}{Initialize $\mathbf w=\{1,\dots,1\}$;}
 \For{\rm $t$ in $i, ..., T$}{
 Randomly sample a class-balanced subset $\mathcal B_{0:i}$ from $\mathcal E_{0:i-1}\cup\mathcal{D}_{i}$,;\\
 Create the local environment $h_i=((\mathcal E_{0:i-1}\cup\mathcal{D}_{i})\setminus\mathcal B_{0:i}, \mathcal B_{0:i})$;\\
 Set the policy $\mathbf p={\mathbf w}/{||\mathbf w||}$;\\
 Sample an action $\mathbf{a}_t\sim\mathbf p$;\\
 \For{\rm $j$ in $i, ..., i+n$}{
 Train $\Theta_{j}$ for $M_1$ epochs by \textbf{Algorithm~\ref{algo_CIL}} with inputs $\Theta_{j-1}$, $\mathbf{a}_t$, $h_i$;\\
 Collect the reward $r_{h_i}(\Theta_j, \mathbf{a}_t)$;}
 Compute the cumulative reward $\hat{R}(\mathbf{a}_t, h_i)$ by Eq.~\ref{eq_reward_1};\\
 Update $\mathbf w$ by Eq.~\ref{eq_exp3_update};\\
 }
 {\color[rgb]{0.165, 0.431, 0.686} \tcp{CIL training}}
 Sample an action $\mathbf{a}_i\sim\mathbf p$;\\
 Train $\Theta_{i}$ for $M_2$ epochs by \textbf{Algorithm~\ref{algo_CIL}} with inputs $\Theta_{i-1}$, $\mathbf{a}_i$, $\mathcal H_i=(\mathcal E_{0:i-1}\cup\mathcal{D}_{i}, \mathcal Q_{0:i})$;\\
 Select new exemplars  $\mathcal E_{0:i}$ from $\mathcal E_{0:i-1}\cup\mathcal{D}_{i}$ using herding~\cite{rebuffi2017icarl}.
 \caption{Our CIL algorithm in Phase $i$ ($i${$\geq$}$1$)}
 \label{algo_Exp3}
 \end{algorithm}

%% file: pseudocode/2_CIL.tex
\begin{algorithm}[ht!]
 \DontPrintSemicolon
 \SetKwInOut{Input}{Input}
 \SetKwInOut{Output}{Output}
 \Input{Old model $\Theta_{\text{old}}$, action $\mathbf{a}=\{\beta, \gamma, \delta, \lambda\}$, environment $h=\{\mathcal T,\mathcal Q\}$.}
 \Output{New model $\Theta$, reward $r_h(\Theta, \mathbf{a})$  (i.e., the validation accuracy).}
 Initialize $\Theta$ with $\Theta_{\text{old}}$;\\
 \For{epochs}{
  \For{mini-batch {\rm ($x$, $y$)$\in\mathcal T$}}{
  Compute the loss $\mathcal{L}_{\text{overall}}(x, y; \Theta_{\text{old}}, \beta, \gamma)$ by Eq.~\ref{eq_overall_loss};\\
  Update $\Theta$ by Eq.~\ref{eq_gd_lr} with $\lambda$;\\
  }
 }
 \For{mini-batch {\rm ($x_{\text{val}}$, $y_{\text{val}}$)$\in\mathcal Q$}}{
 Compute predictions $\mu(x_{\text{val}}; \Theta)$ by Eq.~\ref{eq_pred} with $\delta$;\\ 
 }
 Compute the reward $r_h(\Theta, \mathbf{a})$ using $\{(\mu(x_{\text{val}}),y_{\text{val}})\}_{(x_{\text{val}}, y_{\text{val}})\in\mathcal Q}$.
 \caption{Training and evaluation for action $\mathbf{a}$}
 \label{algo_CIL}
 \end{algorithm}

%% file: sections/5_experiments.tex
\section{Experiments}
\label{sec5_exp}

\input{tables/sota}
\input{tables/sota_imgnet}

We evaluate the proposed method on three CIL benchmarks,  
incorporate our method into three top-performing baseline methods, 
and boost their performances consistently in all settings. 
Below we describe the datasets and implementation details, followed by the results and analyses.

\subsection{Datasets and Implementation Details}
\label{subsec_datasets}

\myparagraph{Datasets.}
We employ CIFAR-100~\cite{krizhevsky2009learning}, ImageNet-Subset~\cite{rebuffi2017icarl} (100 classes), and ImageNet-Full~\cite{russakovsky2015imagenet} (1000 classes) as the benchmarks. We use the same data splits and class orders as the related work~\cite{rebuffi2017icarl,Liu2020AANets,Liu2021RMM}.

\myparagraph{Network architectures.}
We use a modified $32$-layer ResNet~\cite{rebuffi2017icarl} for CIFAR-100 and an $18$-layer ResNet~\cite{He2016ResNet} for ImageNet, following~\cite{rebuffi2017icarl,hou2019lucir,Liu2020AANets}. We deploy the AANets~\cite{Liu2020AANets} for the experiments based on AANets and RMM~~\cite{Liu2021RMM}. Further, we use a cosine normalized classifier without bias terms as the FC classifier, following~\cite{hou2019lucir,liu2020mnemonics}.

\myparagraph{Configurations.} We discretize the hyperparameter search space into $50$ actions, i.e., $\text{card}(\mathbb A)$=$50$. We update the policy for $25$ iterations in each phase, i.e., $T$=$25$. For other configurations, we follow the corresponding baselines. 

\subsection{Results and Analyses}
\label{subsec_results}

Tables~\ref{table_sota} and \ref{table_sota_imgnet} present the results of top-performing baselines \textsl{w/} and \textsl{w/o} our method and some recent related work. Table~\ref{table_ablation} summarizes the results in seven ablative settings. 
Figure~\ref{GradCAM_figure} compares the activation maps (using Grad-CAM~\cite{selvaraju2017gradcam}) produced by diffident methods in TFH and TFS. Figure~\ref{figure_values} shows the values of hyperparameters produced by our method.

\myparagraph{Comparison with the state-of-the-art.} Tables~\ref{table_sota} and \ref{table_sota_imgnet} show that taking our method as a plug-in module for  the state-of-the-art~\cite{Liu2021RMM} and other baselines~\cite{hou2019lucir,Liu2020AANets} consistently improves their performance. For example, RMM~\cite{Liu2021RMM} \textsl{w/} ours gains $4.3$ and $2.2$ percentage points on CIFAR-100 and ImageNet-Full, respectively. Interestingly, we find that we can surpass the baselines more when the number of phases $N$ is larger. E.g., on CIFAR-100, our method improves RMM by $5.7$ percentage points when $N$=$25$, while this number is $2.8$ percentage points when $N$=$5$.
Our explanation is that the forgetting problem is more serious when the number of phases is larger. Thus, we need better hyperparameters to balance stability and plasticity.

\input{tables/ablation}

\myparagraph{Ablation study.} Table~\ref{table_ablation} concerns eight ablation settings, and shows the results in both TFH and TFS for different numbers of phases ($N$=$5$/$25$). The detailed analyses are as follows.

\myparagraph{\emph{1) First block.}} Row~1 shows the baseline~\cite{hou2019lucir}.

\myparagraph{\emph{2) Second block: optimizable hyperparameters.}} In our study, we optimize three kinds of hyperparameters that affect the model's stability and plasticity: KD loss weights $(\beta,\gamma)$, learning rate $\lambda$, and classifier type indicator $\delta$. Comparing Row 2 to Row 1, we can observe that optimizing the KD loss weights boosts the TFS accuracy more significantly. It is because the baseline, LUCIR~\cite{hou2019lucir}, applies a strong regularization term (i.e., feature KD loss) which harms the TFH performance. Our method changes the regularization term by adjusting the KD loss weights, so it achieves better performance. Comparing Row 3 to Row 2, we can see that optimizing the classifier type indicator $\delta$ performs further improvements, especially in TFS. It is because the baseline deploys an FC classifier by default while our method learns to switch between different classifiers in different settings. Comparing Row 4 to Row 3, we can see the effectiveness of optimizing the learning rates in CIL. In summary, it is impressive that optimizing three kinds of hyperparameters together achieves the best results. 

\myparagraph{\emph{3) Third block: hyperparameter learning methods.}}  For Row 5, we use cross-validation (i.e., all past, future, and validation data are accessible) to find a set of fixed hyperparameters and apply them to all phases. We can see that Row 5 results are consistently lower than ours in Row 4, although it can access more data. It shows that we need to update the hyperparameters online in different phases. For Row 6, we use the policy pre-trained in the $0$-th phase of the target CIL task using the framework proposed in \cite{Liu2021RMM} (compared to ours, it is offline). Comparing Row 6 with Row 4, we are happy to see that our online learning algorithm achieves better performance than the offline RL while we use much less training time (see the analysis in the supplementary). {For Row 7, we use the bilevel hyperparameter optimization method~\cite{franceschi2018bilevel}. Comparing Row 7 with Row 4, we can observe that our method achieves more significant performance improvements. The reason is that \cite{franceschi2018bilevel} is designed for time-invariant environments, while our online algorithm can adapt to the time-varying environments in CIL.}.

\input{figures/4_cam}

\myparagraph{Visualizing activation maps.} Figure~\ref{GradCAM_figure} demonstrates the activation maps visualized by Grad-CAM~\cite{selvaraju2017gradcam} for the final model (obtained after five phases) on ImageNet-Subset $5$-phase. The left and right sub-figures show the results for {\emph{training-from-half}} (TFH) and {\emph{training-from-scratch}} (TFS), respectively. We can observe: 1) LUCIR~\cite{hou2019lucir} makes predictions according to foreground (correct) and background (incorrect) regions in the TFH and TFS settings, respectively; 2) iCaRL~\cite{rebuffi2017icarl} behaves opposite to LUCIR in the two settings;  3) our method always makes predictions according to foreground (correct) regions in both TFH and TFS. The reasons are as follows. LUCIR applies a strong (feature) KD by default, so it performs better in TFH. iCaRL applies a weak (logit) KD by default, so it performs better in TFS. Our method can adjust the hyperparameters to change between different KD losses, so it performs well in both settings. 

\input{figures/5_values}

\myparagraph{Hyperparameter values.}  Figure~\ref{figure_values} shows the  hyperparameter values produced by our
policy on CIFAR-100 $25$-phase. 

\myparagraph{\emph{1) KD loss weights $\beta$ and $\gamma$.}}
From the figure, we have two observations. a) The policy learns to produce a larger initial value for $\gamma$ and $\beta$ in TFH and TFS, respectively. Our explanation is as follows. In TFH, we already have a pre-trained model, so we need a strong regularization term (i.e., feature KD loss) to make the model more stable and avoid forgetting. In TFS, we start from random initialization, so we need a weak regularization term (i.e., logit KD loss) to improve the model's plasticity. b) Both  $\beta$ and $\gamma$ increase in TFH, while $\beta$ decreases and $\gamma$ increases in TFS. It is because we need stronger regularization to maintain the knowledge when more data is observed. The policy achieves that in different ways: it assigns higher weights for both KD losses in TFH, while it transfers from logit KD to feature KD in TFS.

\myparagraph{\emph{2) Learning rates $\lambda$.}}
In Figure~\ref{figure_values}, we can observe that the learning rate in TFS is much higher than in TFH. It can be explained that 1) we need a higher learning rate in TFS as the model is trained from scratch and needs to capture more new knowledge; 2)~we need a lower learning rate in TFH because we need to avoid forgetting the pre-trained model.

\myparagraph{\emph{3) Classifier type indicator $\delta$.}} On CIFAR-100 $25$-phase, our policy learned to choose the NCM and FC classifiers in TFS and TFH, respectively. 

%% file: tables/sota.tex
\newcommand{\redtf}[1]{\textcolor[rgb]{0.8, 0.2, 0.06}{#1}}
\begin{table*}[t]
\small
\vspace{0.1cm}
\begin{center}
{\renewcommand\arraystretch{0.9}
\setlength{\tabcolsep}{1.27mm}{
\begin{tabular}{lccccccccccccccc}
\toprule
\multirow{2.5}{*}{\textbf{Methods}} &\multicolumn{3}{c}{\textbf{CIFAR-100}, $N$=5}&& \multicolumn{3}{c}{\textbf{CIFAR-100}, $N$=25}&&\multicolumn{3}{c}{\textbf{ImageNet-Subset}, $N$=5}&& \multicolumn{3}{c}{\textbf{ImageNet-Subset}, $N$=25}\\
\cmidrule{2-4} \cmidrule{6-8}  \cmidrule{10-12} \cmidrule{14-16}
& TFH & TFS & Avg.&& TFH & TFS & Avg.&&TFH & TFS & Avg.&& TFH & TFS & Avg.\\
\midrule
iCaRL~\cite{rebuffi2017icarl}& 58.1 & 64.0 & 61.0 && 48.1 & 53.2 & 50.7 && 65.3 & 70.4 & 67.9 && 53.0 & 53.5 & 53.3\\
PODNet~\cite{douillard2020podnet} &  64.7 & 63.6 & 64.2 && 60.3 & 45.3 & 52.8 && 64.3 & 58.9 & 61.6 && 68.3 & 39.1 & 53.7\\
DER~\cite{yan2021DER} &  67.6 & {72.3} & 70.0 && 65.5 & {67.3} & 66.4 && 78.4 & 76.9 & 77.7 && 75.4 & 71.0 & 73.2\\
FOSTER~\cite{wang2022foster}&{70.4}&72.5&71.5&&63.8&\textbf{70.7}&67.3&&80.2&78.3&79.3&&69.3&72.9&71.1\\
\midrule
LUCIR~\cite{hou2019lucir}& 63.1\tiny{$\pm$0.7} & 63.0\tiny{$\pm$0.6} & 63.1\tiny{$\pm$0.7} && 57.5\tiny{$\pm$0.4} & 49.2\tiny{$\pm$0.5} & 53.4\tiny{$\pm$0.5} && 65.3\tiny{$\pm$0.6} & 66.7\tiny{$\pm$0.5} & 66.0\tiny{$\pm$0.6} && 61.4\tiny{$\pm$0.7} & 46.2\tiny{$\pm$0.8} & 53.8\tiny{$\pm$0.8}\\
 \cellcolor{mygray-bg}{\ \ \textsl{w/} ours}& \cellcolor{mygray-bg}{63.9\tiny{$\pm$0.6}} & \cellcolor{mygray-bg}{64.9\tiny{$\pm$0.5}} & \cellcolor{mygray-bg}{64.4\tiny{$\pm$0.6}} &\cellcolor{mygray-bg}{}& \cellcolor{mygray-bg}{59.3\tiny{$\pm$0.5}} & \cellcolor{mygray-bg}{52.4\tiny{$\pm$0.5}} & \cellcolor{mygray-bg}{55.9\tiny{$\pm$0.5}} &\cellcolor{mygray-bg}{}& \cellcolor{mygray-bg}{70.6\tiny{$\pm$0.7}} & \cellcolor{mygray-bg}{68.4\tiny{$\pm$0.6}} & \cellcolor{mygray-bg}{69.5\tiny{$\pm$0.7}} &\cellcolor{mygray-bg}{}& \cellcolor{mygray-bg}{62.9\tiny{$\pm$0.6}} & \cellcolor{mygray-bg}{54.1\tiny{$\pm$0.6}} & \cellcolor{mygray-bg}{58.5\tiny{$\pm$0.6}}\\
 \\[-11pt]
 \cellcolor{mygray-bg}{}& \cellcolor{mygray-bg}{\redtf{\tiny{$\uparrow$0.8}}} & \cellcolor{mygray-bg}{\redtf{\tiny{$\uparrow$1.9}}} & \cellcolor{mygray-bg}{\redtf{\tiny{$\uparrow$1.3}}} &\cellcolor{mygray-bg}{}& \cellcolor{mygray-bg}{\redtf{\tiny{$\uparrow$1.8}}} & \cellcolor{mygray-bg}{\redtf{\tiny{$\uparrow$3.2}}} & \cellcolor{mygray-bg}{\redtf{\tiny{$\uparrow$2.5}}} &\cellcolor{mygray-bg}{}& \cellcolor{mygray-bg}{\redtf{\tiny{$\uparrow$5.3}}} & \cellcolor{mygray-bg}{\redtf{\tiny{$\uparrow$1.7}}} & \cellcolor{mygray-bg}{\redtf{\tiny{$\uparrow$3.5}}} &\cellcolor{mygray-bg}{}& \cellcolor{mygray-bg}{\redtf{\tiny{$\uparrow$1.5}}} & \cellcolor{mygray-bg}{\redtf{\tiny{$\uparrow$7.9}}} & \cellcolor{mygray-bg}{\redtf{\tiny{$\uparrow$4.7}}}\\
 \\[-12pt]
 \midrule
AANets~\cite{Liu2020AANets}& 65.3\tiny{$\pm$0.4} & 63.1\tiny{$\pm$0.3} & 64.2\tiny{$\pm$0.4} && 63.2\tiny{$\pm$0.3} & 44.4\tiny{$\pm$0.4} & 53.8\tiny{$\pm$0.4} && 77.0\tiny{$\pm$0.7} & 68.9\tiny{$\pm$0.6} & 73.0\tiny{$\pm$0.7} && 72.2\tiny{$\pm$0.6} & 60.7\tiny{$\pm$0.5} & 66.5\tiny{$\pm$0.6}\\
 \cellcolor{mygray-bg}{\ \ \textsl{w/} ours}&  \cellcolor{mygray-bg}{67.0\tiny{$\pm$0.3}} & \cellcolor{mygray-bg}{65.1\tiny{$\pm$0.3}} & \cellcolor{mygray-bg}{66.1\tiny{$\pm$0.3}} &\cellcolor{mygray-bg}{}& \cellcolor{mygray-bg}{64.1\tiny{$\pm$0.4}} & \cellcolor{mygray-bg}{50.3\tiny{$\pm$0.5}} & \cellcolor{mygray-bg}{57.2\tiny{$\pm$0.5}} &\cellcolor{mygray-bg}{}& \cellcolor{mygray-bg}{77.3\tiny{$\pm$0.6}} & \cellcolor{mygray-bg}{70.6\tiny{$\pm$0.5}} & \cellcolor{mygray-bg}{74.0\tiny{$\pm$0.6}} &\cellcolor{mygray-bg}{}& \cellcolor{mygray-bg}{72.9\tiny{$\pm$0.5}} & \cellcolor{mygray-bg}{64.8\tiny{$\pm$0.5}} & \cellcolor{mygray-bg}{68.9\tiny{$\pm$0.5}}\\
 \\[-11pt]
 \cellcolor{mygray-bg}{}& \cellcolor{mygray-bg}{\redtf{\tiny{$\uparrow$1.7}}} & \cellcolor{mygray-bg}{\redtf{\tiny{$\uparrow$2.0}}} & \cellcolor{mygray-bg}{\redtf{\tiny{$\uparrow$1.9}}} &\cellcolor{mygray-bg}{}& \cellcolor{mygray-bg}{\redtf{\tiny{$\uparrow$0.9}}} & \cellcolor{mygray-bg}{\redtf{\tiny{$\uparrow$5.9}}} & \cellcolor{mygray-bg}{\redtf{\tiny{$\uparrow$3.4}}} &\cellcolor{mygray-bg}{}& \cellcolor{mygray-bg}{\redtf{\tiny{$\uparrow$0.3}}} & \cellcolor{mygray-bg}{\redtf{\tiny{$\uparrow$1.7}}} & \cellcolor{mygray-bg}{\redtf{\tiny{$\uparrow$1.0}}} &\cellcolor{mygray-bg}{}& \cellcolor{mygray-bg}{\redtf{\tiny{$\uparrow$0.7}}} & \cellcolor{mygray-bg}{\redtf{\tiny{$\uparrow$4.1}}} & \cellcolor{mygray-bg}{\redtf{\tiny{$\uparrow$2.4}}}\\
 \\[-12pt]
 \midrule
RMM~\cite{Liu2021RMM}& 67.6\tiny{$\pm$0.7} & 70.4\tiny{$\pm$0.8} & 69.0\tiny{$\pm$0.8} && 65.6\tiny{$\pm$0.6} & 58.4\tiny{$\pm$0.6} & 62.0\tiny{$\pm$0.6} && 79.5\tiny{$\pm$0.2} & 80.5\tiny{$\pm$0.3} & 80.0\tiny{$\pm$0.3} && 75.0\tiny{$\pm$0.3} & 71.6\tiny{$\pm$0.3} & 73.3\tiny{$\pm$0.3}\\
 \cellcolor{mygray-bg}{\ \ \textsl{w/} ours}&  \cellcolor{mygray-bg}{\textbf{70.8}\tiny{$\pm$0.7}} & \cellcolor{mygray-bg}{\textbf{72.7}\tiny{$\pm$0.6}} & \cellcolor{mygray-bg}{\textbf{71.8}\tiny{$\pm$0.7}} &\cellcolor{mygray-bg}{}& \cellcolor{mygray-bg}{\textbf{69.5}\tiny{$\pm$0.8}} & \cellcolor{mygray-bg}{{65.9}\tiny{$\pm$0.7}} & \cellcolor{mygray-bg}{\textbf{67.7}\tiny{$\pm$0.8}} &\cellcolor{mygray-bg}{}& \cellcolor{mygray-bg}{\textbf{81.0}\tiny{$\pm$0.3}} & \cellcolor{mygray-bg}{\textbf{82.2}\tiny{$\pm$0.4}} & \cellcolor{mygray-bg}{\textbf{81.6}\tiny{$\pm$0.4}} &\cellcolor{mygray-bg}{}& \cellcolor{mygray-bg}{\textbf{76.1}\tiny{$\pm$0.2}} & \cellcolor{mygray-bg}{\textbf{73.2}\tiny{$\pm$0.4}} & \cellcolor{mygray-bg}{\textbf{74.7}\tiny{$\pm$0.3}}\\
  \\[-11pt]
 \cellcolor{mygray-bg}{}& \cellcolor{mygray-bg}{\redtf{\tiny{$\uparrow$3.2}}} & \cellcolor{mygray-bg}{\redtf{\tiny{$\uparrow$2.3}}} & \cellcolor{mygray-bg}{\redtf{\tiny{$\uparrow$2.8}}} &\cellcolor{mygray-bg}{}& \cellcolor{mygray-bg}{\redtf{\tiny{$\uparrow$3.9}}} & \cellcolor{mygray-bg}{\redtf{\tiny{$\uparrow$7.5}}} & \cellcolor{mygray-bg}{\redtf{\tiny{$\uparrow$5.7}}} &\cellcolor{mygray-bg}{}& \cellcolor{mygray-bg}{\redtf{\tiny{$\uparrow$1.5}}} & \cellcolor{mygray-bg}{\redtf{\tiny{$\uparrow$1.7}}} & \cellcolor{mygray-bg}{\redtf{\tiny{$\uparrow$1.6}}} &\cellcolor{mygray-bg}{}& \cellcolor{mygray-bg}{\redtf{\tiny{$\uparrow$1.1}}} & \cellcolor{mygray-bg}{\redtf{\tiny{$\uparrow$1.6}}} & \cellcolor{mygray-bg}{\redtf{\tiny{$\uparrow$1.4}}}\\
 \\[-12pt]
\bottomrule
\end{tabular}}
}
\end{center}
\vspace{-0.5cm}
\caption{Average accuracy (\%) across all phases on CIFAR-100 and ImageNet-Subset. The first block shows some recent CIL methods.
The second block shows three top-performing baselines~\cite{hou2019lucir,Liu2020AANets,Liu2021RMM} \textsl{w/} and \textsl{w/o} our method plugged in. ``TFH'' and  ``TFS'' denote the {\emph{training-from-half}} and {\emph{training-from-scratch}} settings, respectively. ``Avg.'' shows the average of the ``TFH'' and  ``TFS'' results. For ``AANets''~\cite{Liu2020AANets}, we use its version based on PODNet~\cite{douillard2020podnet}. We rerun the baselines
using their open-source code in a unified setting for a fair comparison. 
} 
\label{table_sota}
\end{table*}

%% file: tables/sota_imgnet.tex
\begin{table}
\centering
{\small
\renewcommand\arraystretch{0.9}
\setlength{\tabcolsep}{3.1mm}{
\begin{tabular}{lccccccccccccccc}
\toprule
\multirow{2.5}{*}{\textbf{Methods}} &\multicolumn{3}{c}{\textbf{ImageNet-Full}, $N$=5}\\
\cmidrule{2-4} 
& TFH & TFS & Avg.\\
\midrule
LUCIR~\cite{hou2019lucir}& 64.5\tiny{$\pm$0.3} & 62.7\tiny{$\pm$0.4} & 62.0\tiny{$\pm$0.4} \\
 \cellcolor{mygray-bg}{\ \ \textsl{w/} ours}& \cellcolor{mygray-bg}{65.8\tiny{$\pm$0.3} \ \redtf{\tiny{$\uparrow$1.3}}} & \cellcolor{mygray-bg}{66.1\tiny{$\pm$0.3} \ \redtf{\tiny{$\uparrow$3.4}}} & \cellcolor{mygray-bg}{66.0\tiny{$\pm$0.3} \ \redtf{\tiny{$\uparrow$2.4}}} \\\
 \\[-12pt]
 \midrule
RMM~\cite{Liu2021RMM}& 69.0\tiny{$\pm$0.5} & 66.1\tiny{$\pm$0.4} & 67.6\tiny{$\pm$0.5} \\
 \cellcolor{mygray-bg}{\ \ \textsl{w/} ours}& \cellcolor{mygray-bg}{\textbf{70.7}\tiny{$\pm$0.5} \ \redtf{\tiny{$\uparrow$1.7}}} & \cellcolor{mygray-bg}{\textbf{68.9}\tiny{$\pm$0.5} \ \redtf{\tiny{$\uparrow$2.8}}} & \cellcolor{mygray-bg}{\textbf{69.8}\tiny{$\pm$0.5} \ \redtf{\tiny{$\uparrow$2.2}}} \\
  \\[-12pt]
\bottomrule
\end{tabular}}}
\caption{Average accuracy (\%) on ImageNet-Full.} 
\vspace{-0.4cm}
\label{table_sota_imgnet}
\end{table}

%% file: tables/ablation.tex
\begin{table}[t]
\small
\begin{center}
{\renewcommand\arraystretch{0.9}
\setlength{\tabcolsep}{1.45mm}{
\begin{tabular}{lccccccccccccccc}
\toprule
\multirow{2.5}{*}{{No.}}  & 
\multicolumn{3}{c}{Optimizing}&& \multicolumn{2}{c}{$N$=5 } && \multicolumn{2}{c}{$N$=25 }\\
\cmidrule{2-4} \cmidrule{6-7} \cmidrule{9-10} 
& ($\beta,\gamma$) & {\textcolor{white}{(a}}$\delta${\textcolor{white}{a,)}} & {\textcolor{white}{(a}}$\lambda${\textcolor{white}{a)}} && TFH & TFS && TFH & TFS \\
\midrule
1  &\multicolumn{3}{c}{Baseline}&& 63.11 & 62.96  && 57.47 & 49.16 \\
\midrule
2   &\checkmark&&&& 63.20 & 63.60  && 58.27 & 50.91 \\
3  &\checkmark&\checkmark&&& 63.23 & 64.08  && 58.20 & 51.94 \\
{4 }  &{\checkmark}&{\checkmark}&{\checkmark}&{}& {\textbf{63.88}} & {\textbf{64.92}} &{}& {\textbf{59.27}} & {\textbf{52.44}} \\
\midrule
5 & \multicolumn{3}{c}{Cross-val fixed} && 63.33 & 64.02 & & 57.50 & 51.64 \\
6 & \multicolumn{3}{c}{Offline RL ~\cite{Liu2021RMM}} && 63.42 & 63.88 & & 58.12 & 51.53  \\
7 & \multicolumn{3}{c}{Bilevel HO~\cite{franceschi2018bilevel}} && 63.20 & 63.02 & & 57.56 & 49.42  \\
\bottomrule
\end{tabular}}
}
\end{center}
\vspace{-3mm}
\caption{Ablation results (average accuracy \%) on CIFAR-100. ($\beta,\gamma$) are KD loss weights. $\lambda$ and $\delta$ denote learning rates and classifier types, respectively.  The baseline is LUCIR~\cite{hou2019lucir}. Row 4 shows our best result. 
 } 
\vspace{-0.2cm}
\label{table_ablation}
\end{table}

%% file: figures/4_cam.tex
\begin{figure}[t]
\centering
\includegraphics[width=0.48\textwidth]{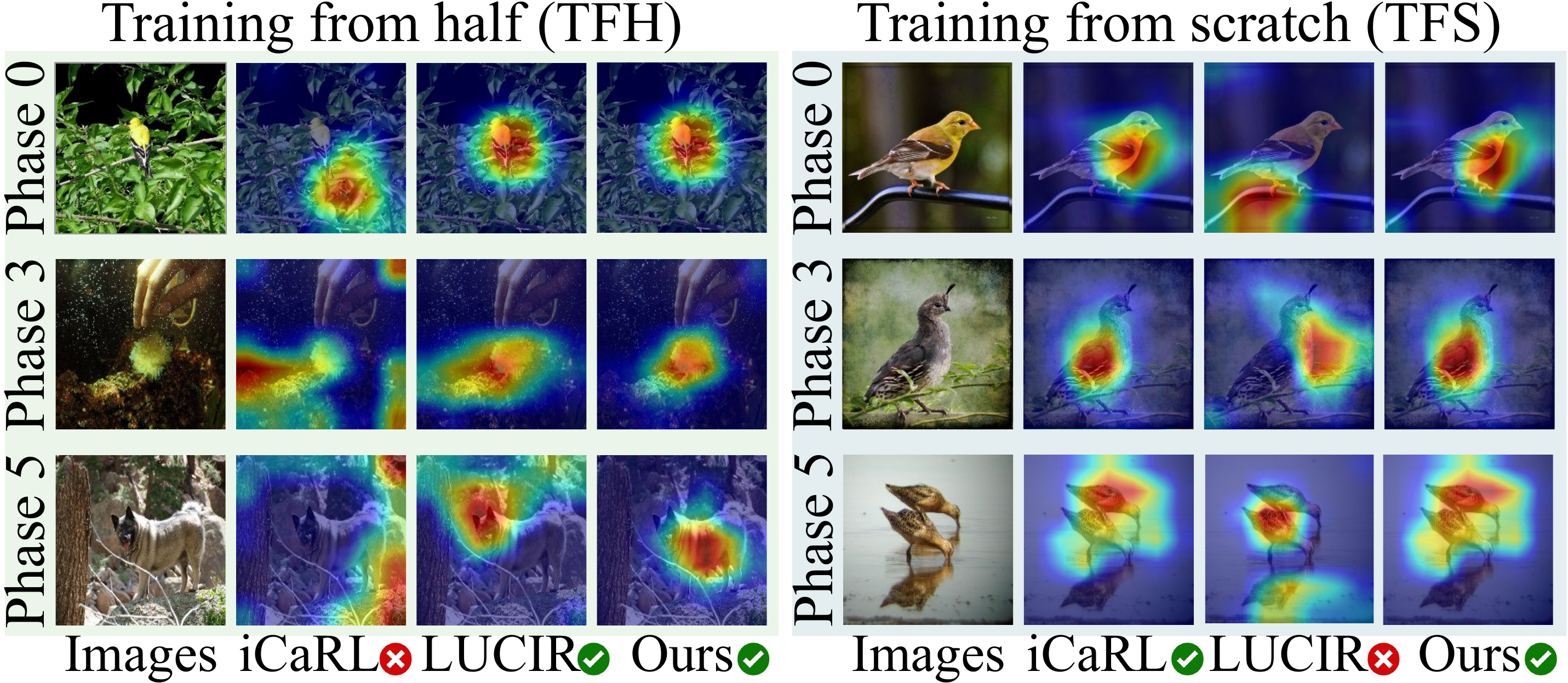}
\caption{The activation maps using Grad-CAM~\cite{selvaraju2017gradcam} for the last phase model on ImageNet-Subset $5$-phase. Samples are
selected from the classes coming in the $0$-th, $3$-rd, and $5$-th phases. Green ticks mean successful activation of discriminative features on object regions, while red crosses mean unsuccessful.}
  \label{GradCAM_figure}
  \vspace{-0.1cm}
\end{figure}

%% file: figures/5_values.tex
\begin{figure}
\begin{center}
\includegraphics[height=0.95in]{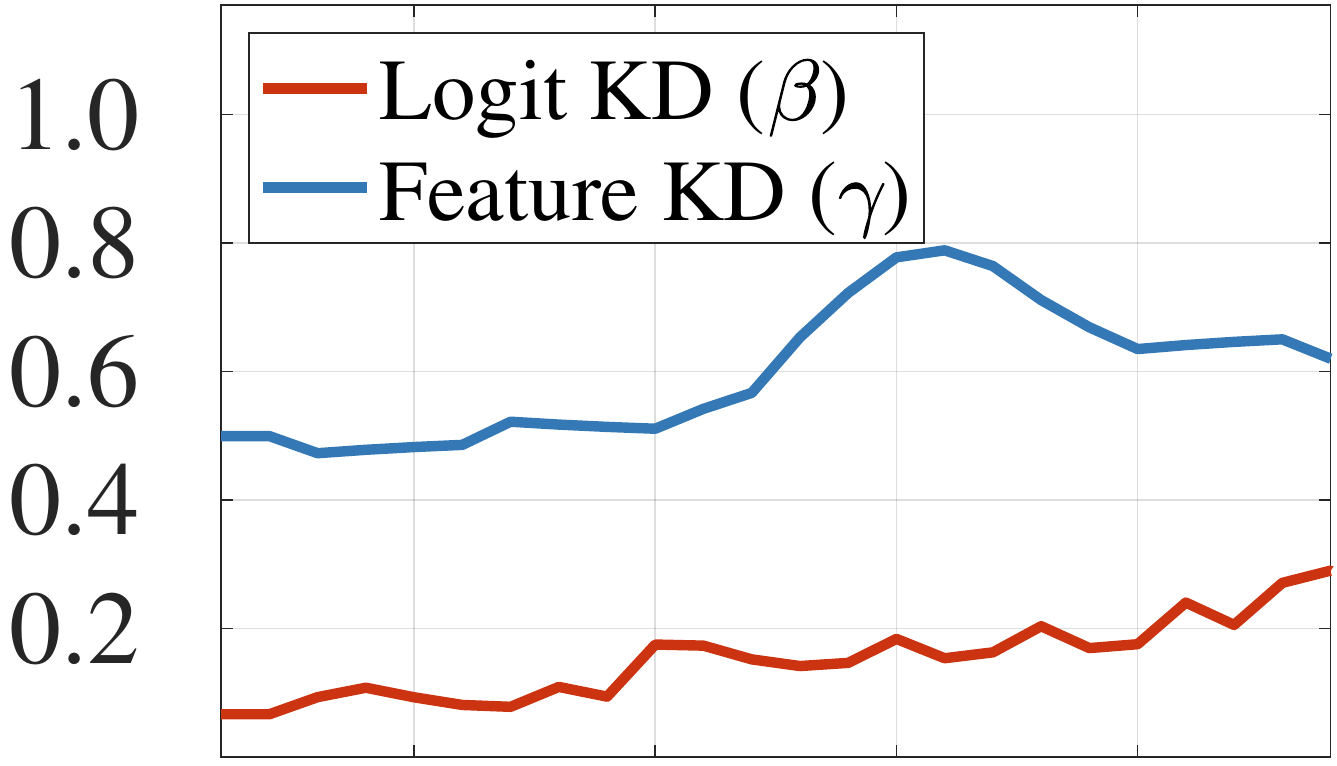}
\hspace{0.1cm}
\includegraphics[height=0.95in]{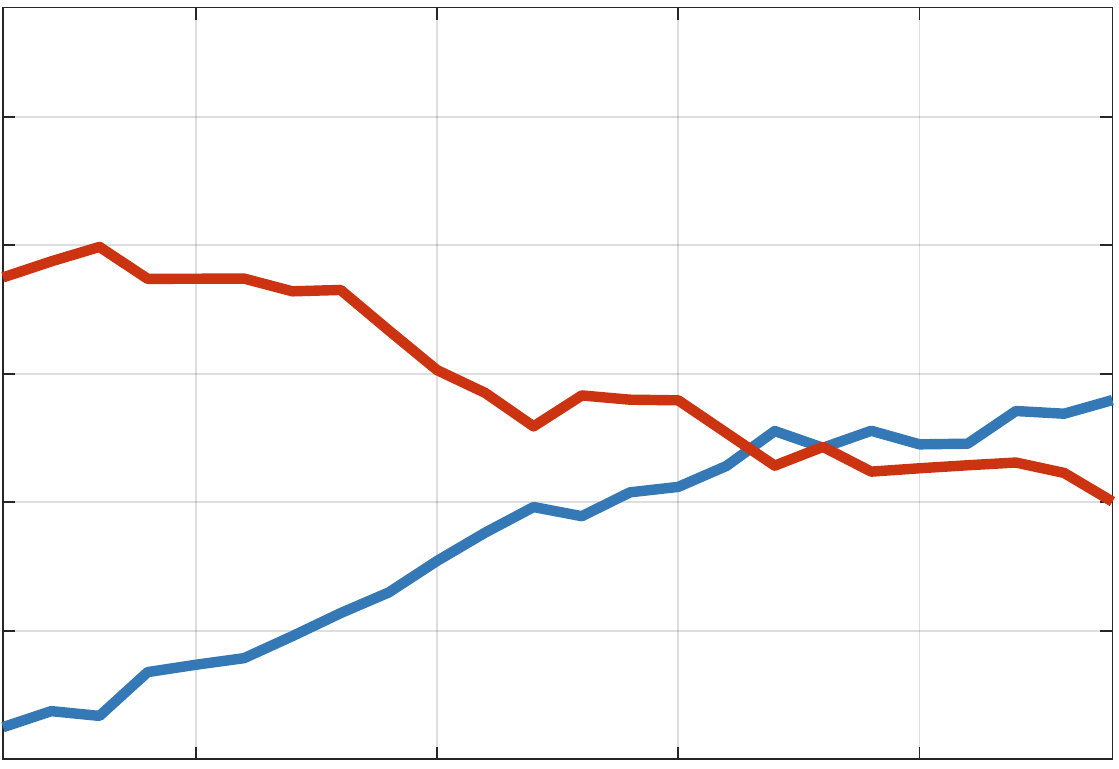}
\includegraphics[height=1.26in]{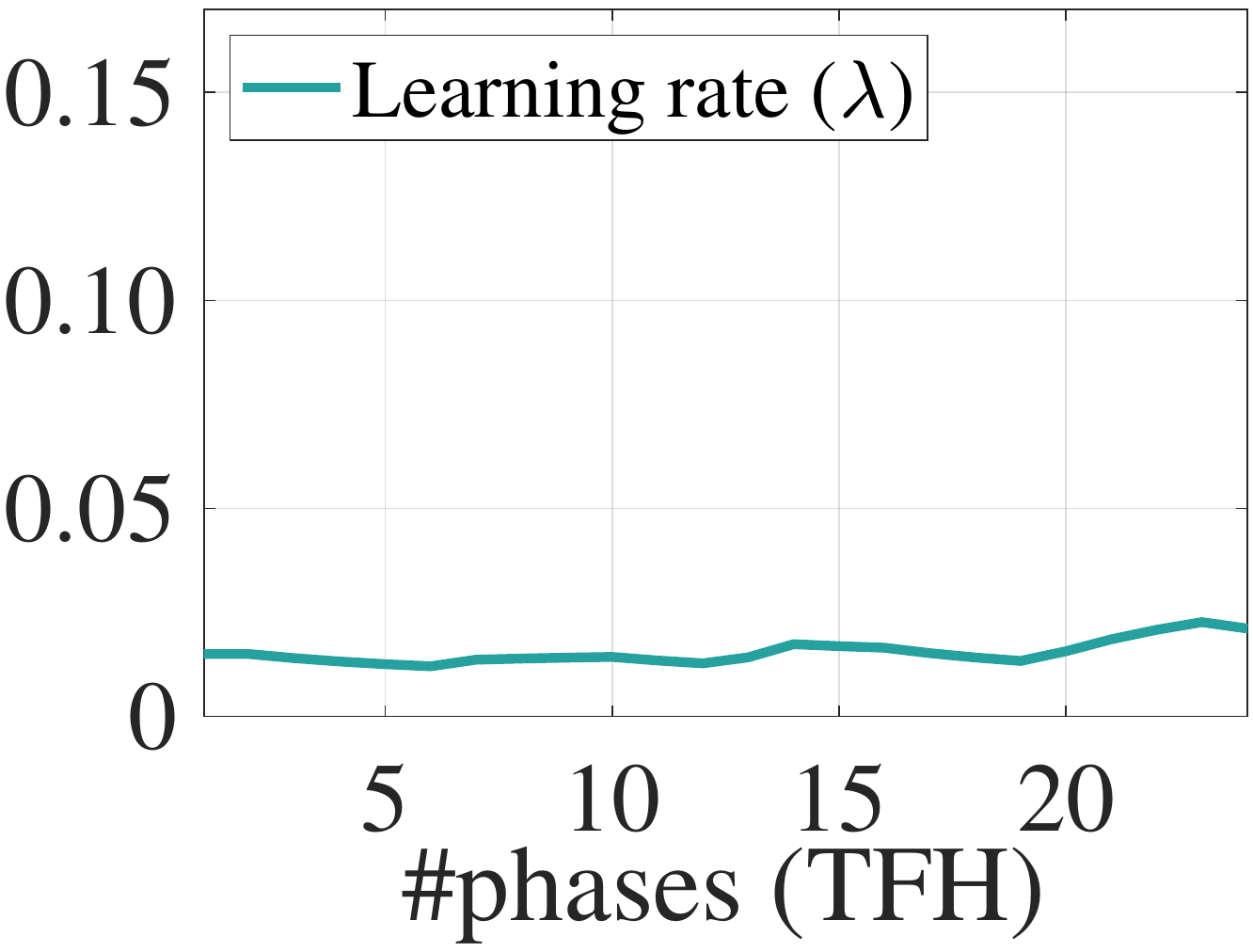}
\hspace{0.1cm}
\includegraphics[height=1.24in]{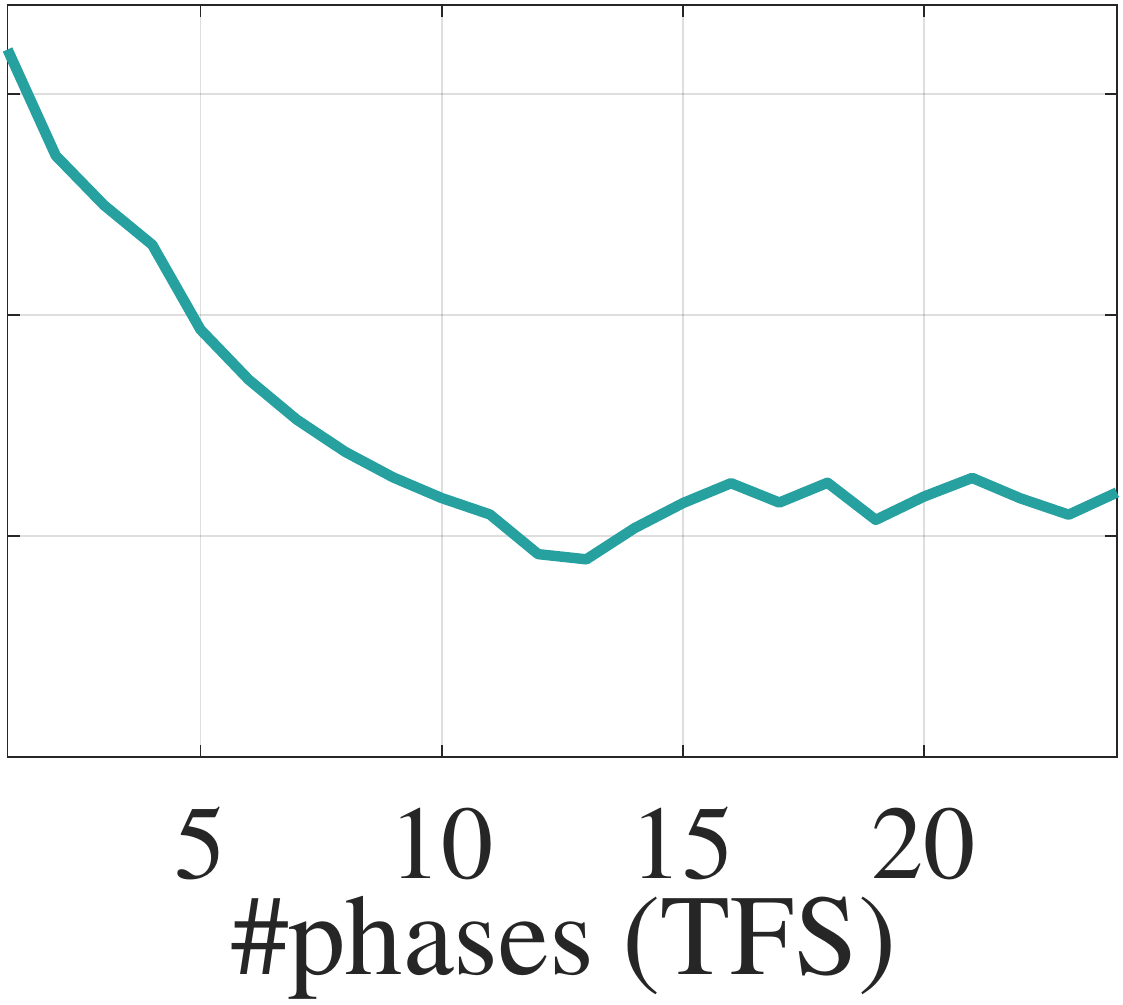}
\end{center}
  \vspace{-0.4cm}
\caption{The hyperparameter values produced by our policy on CIFAR-100 $25$-phase for LUCIR~\cite{hou2019lucir} \textsl{w/} ours. We smooth all curves with a rate of $0.8$ for better visualization. 
}
\vspace{-0.2cm}
\label{figure_values}
\end{figure}

%% file: sections/6_conclusion.tex
\section{Conclusions}

In this study, we introduce a novel framework that allows us to optimize hyperparameters online to balance the stability and plasticity in CIL. To achieve this, we formulate the CIL task as an online MDP and learn a policy to produce the hyperparameters. Our approach is generic, and it can be easily applied to existing methods to achieve large-margin improvements in both TFS and TFH settings. It is worth mentioning that our method can also be applied to optimize other key hyperparameters in CIL.

\section*{Acknowledgments}

This research was supported by A*STAR under its AME YIRG Grant (Project No. A20E6c0101).

%% file: sections/7_supplementary.tex
\newcommand{\beginsupp}{%
        \setcounter{table}{0}
        \renewcommand{\thetable}{S\arabic{table}}%
        \setcounter{figure}{0}
        \renewcommand{\thefigure}{S\arabic{figure}}%
     }

\newcommand{\myparagraphsupp}[1]{\vspace{0.1em}\noindent{\textcolor[rgb]{0.93,0.47,0.32}{#1}}}

\beginsupp
\setcounter{section}{0}
\renewcommand\thesection{\Alph{section}}
\noindent

\noindent
{\Large {\textbf{Supplementary materials}}}

\vspace{0.5cm}

\section{Training Time}

\myparagraphsupp{Supplementary to Section~\ref{subsec_results}~\textbf{Results and Analyses}.}


Table~\ref{table_ablation_time} shows the training time for different policy learning methods. We can see that the Offline RL method based on pseudo tasks~\cite{Liu2021RMM} requires around $400$ times of additional training time (over the baseline). While our online learning method ($1.4$ times when updating the policy every $5$ phase) is much faster.

We can speed up the training process of our method by reducing the policy updating frequency because it is intuitive to use similar hyperparameters in adjacent phases. We provide the results of updating the policy every $2$ and $5$ phase in Table~\ref{table_ablation_supp} (Rows 2 and 3), respectively. Comparing Row 3 to Row 1, we can still improve the baseline by around $2.2$ percentage points when updating the policy every two phases. So it is practical to reduce the policy updating frequency when we need to speed up the training.

\input{tables/ablation_supp}

\input{tables/time}

\section{Discussions}

\myparagraph{Limited number of policy parameters.}  We deliberately choose a relatively small parameter dimension to reduce computation costs in each phase. We would like to highlight that our goal is not to optimize the hyperparameters continuously--also impossible due to limited information on the future data and limited computational resources in CIL settings. Our goal is to learn to select the hyperparameters under online limited information. We aim to use limited computation to achieve better performance compared to the works using fixed hyperparameters~\cite{douillard2020podnet,Li18LWF,yan2021DER,rebuffi2017icarl}. Our empirical results demonstrate improved performance even with a relatively small policy parameter number, e.g., boosting RMM~\cite{Liu2021RMM} by $5.7$ percentage points using $150$ policy parameters (CIFAR-100, $N$=$25$, Table~\ref{table_sota}).
Further, we would like to mention that we can easily increase the number of policy parameters in our algorithm if we have more computational resources. We expect 
it can achieve further improvements.

\myparagraph{Dependency on previous models and actions.}  We are fully aware of the dependency among phases, and we formulate the CIL problem as an online MDP to capture this dependency. For our algorithm design, we approximately decouple  online MDP as a series of online optimization/multi-armed bandit problems. Such a decoupling follows a classical methodology in the online MDP literature and requires the following assumptions. 1) Fast mixing: in CIL, the hyperparameters in an early phase do not have much impact on the test accuracy of the classes observed in the current phase.
2) The algorithm changes the hyperparameters slowly. We can observe this in Figure~\ref{figure_values}. Thus, these assumptions fit the CIL problem.

\section{Datasets}

\myparagraphsupp{Supplementary to Section~\ref{subsec_datasets} \textbf{Datasets and Implementation Details}.}

We use three benchmarks based on two datasets, CIFAR-100~\cite{krizhevsky2009learning} and ImageNet~\cite{russakovsky2015imagenet} and follow the same data splits in related work~\cite{douillard2020podnet,rebuffi2017icarl,Liu2020AANets,Liu2021RMM}. There are two CIL benchmarks on the ImageNet: ImageNet-Subset using a subset of $100$ classes, and ImageNet-Full using the full set of $1,000$ classes. We use exactly the same class order as the baselines~\cite{hou2019lucir,Liu2020AANets,Liu2021RMM} for fair comparison.

\section{Network Architectures Details}

\myparagraphsupp{Supplementary to Section~\ref{subsec_datasets} \textbf{Datasets and Implementation Details}.}

\myparagraph{CIFAR-100.} We deploy a $32$-layer ResNet for CIFAR-100, following~\cite{rebuffi2017icarl,hou2019lucir,liu2020mnemonics,Liu2020AANets,douillard2020podnet}.
This ResNet consists of $1$ initial convolution layer and $3$ residual blocks (in a single branch). Each block has $10$ convolution layers with $3\times3$ kernels. The number of filters starts from $16$ and is doubled every next block. After these $3$ blocks, there is an average-pooling layer to compress the output feature maps to a feature embedding. 

\myparagraph{ImageNet-Subset and ImageNet-Full.} We deploy a standard ResNet-18 for the experiments on ImageNet-Subset and ImageNet-Full, following~\cite{rebuffi2017icarl,hou2019lucir,liu2020mnemonics,Liu2020AANets,douillard2020podnet}. We use a relatively shallow network for ImageNet-Subset and ImageNet-Full because this is the standard setting in the related work and can speed up the experiments. 

\myparagraph{AANets.} We deploy POD-AANets~\cite{Liu2020AANets,douillard2020podnet} when the baselines are AANets~\cite{Liu2020AANets} and RMM~\cite{Liu2021RMM}, following the original papers. On CIFAR-100, we convert these $3$ blocks in the ResNet-32 into three levels of blocks and each level consists of a stable block and a plastic block.
Similarly, we build AANets for ImageNet benchmarks but take an $18$-layer ResNet~\cite{He2016ResNet} as the baseline architecture~\cite{hou2019lucir,liu2020mnemonics}. 
Please note that there is no architecture change applied to the classifiers, i.e., using the same FC layers as in~\cite{hou2019lucir,liu2020mnemonics}.

\section{More Implementation Details}
\label{supp_sec_implementation}

\myparagraphsupp{Supplementary to Section~\ref{subsec_datasets} \textbf{Datasets and Implementation Details}.}

{\myparagraph{CIL training configuration.}}
For fair comparison, our training configuration is almost the same as in~\cite{douillard2020podnet,hou2019lucir,Liu2021RMM,Liu2020AANets}.
Specifically, we train $160$ ($90$) epochs in each phase on CIFAR-100 (ImageNet), i.e., $M_2$=$160$ for CIFAR-100 and $M_2$=$90$ for ImageNet. We use an SGD optimizer with the momentum $0.9$ and the batch size $128$ to train the models in all settings.

\myparagraph{LUCIR \textsl{w/} ours.} Our overall KD loss consists of two different KD losses when the baseline is LUCIR~\cite{hou2019lucir}. The two  KD losses are logit KD loss~\cite{Hinton15KnowledgeDistillation} and feature KD loss~\cite{hou2019lucir}, respectively. We implement the logit KD loss following~\cite{rebuffi2017icarl}. Further, we apply the same implementation as the Less-Forget Constraint in~\cite{hou2019lucir}. Besides, we exactly follow the official open-source code of LUCIR~\cite{hou2019lucir} for other implementation details.

\myparagraph{AANets \textsl{w/} ours.} For ``AANets''~\cite{Liu2020AANets}, we use its version based on PODNet~\cite{douillard2020podnet}, i.e., POD-AANets. We incorporate three KD losses in this framework:  logit KD loss~\cite{Hinton15KnowledgeDistillation}, feature KD loss~\cite{hou2019lucir}, and POD loss~\cite{douillard2020podnet}. Thus, we learn a policy to generate three different weights for different KD losses. Furthermore, we exactly follow the official open-source code of POD-AANets~\cite{Liu2020AANets} for other implementation details.

\myparagraph{RMM \textsl{w/} ours.} We use POD-AANets \textsl{w/} RMM~\cite{Liu2021RMM} as the baseline. Therefore, we also optimize the KD loss weights for three KD losses:  logit KD loss~\cite{Hinton15KnowledgeDistillation}, feature KD loss~\cite{hou2019lucir}, and POD loss~\cite{douillard2020podnet}. In each phase, we apply the same total memory budget as the other two baselines~\cite{hou2019lucir,Liu2020AANets}. Then, we apply RMM to adjust the memory allocation. We exactly follow the official open-source code of RMM~\cite{Liu2021RMM} for other implementation details.


\section{Hardware Information}
\label{supp_sec_add_info}

We run experiments using GPU workstations as follows,
\begin{itemize}
    \item \textbf{CPU}: AMD EPYC 7502P 32-Core Processor
    \item \textbf{GPU}: NVIDIA Quadro RTX 8000, 48 GB GDDR6
    \item \textbf{Memory}: 1024 GiB, DDR4, 3200 MHz, ECC
\end{itemize}

\section{License Information}
\label{supp_sec_license}

\myparagraph{Code licenses.} The code for the following papers using the MIT License: AANets~\cite{Liu2020AANets}, iCaRL~\cite{rebuffi2017icarl}, PODNet~\cite{douillard2020podnet}, DER~\cite{yan2021DER}, and RMM~\cite{Liu2021RMM}.

\myparagraph{Datasets.} We use two datasets in our paper: CIFAR-100~\cite{krizhevsky2009learning} and ImageNet~\cite{russakovsky2015imagenet}. The data for both datasets are downloaded from their official websites and allowed to use for non-commercial research and educational purposes.

%% file: tables/ablation_supp.tex
\begin{table}[ht!]
\small
\begin{center}
{\renewcommand\arraystretch{1.0}
\setlength{\tabcolsep}{1.2mm}{
\begin{tabular}{lccccccccccccccc}
\toprule
\multirow{2.5}{*}{{No.}}  & 
\multicolumn{3}{c}{Optimizing}&& \multicolumn{2}{c}{$N$=5 (acc. \%)} && \multicolumn{2}{c}{$N$=25 (acc. \%)}\\
\cmidrule{2-4} \cmidrule{6-7} \cmidrule{9-10} 
& ($\beta,\gamma$) & {\textcolor{white}{(a}}$\delta${\textcolor{white}{a,)}} & {\textcolor{white}{(a}}$\lambda${\textcolor{white}{a)}} && TFH & TFS && TFH & TFS \\
\midrule
1  &\multicolumn{3}{c}{Baseline}&& 63.11 & 62.96  && 57.47 & 49.16 \\
\midrule
2 & \multicolumn{3}{c}{Every $2$ phases} && 63.57 & 64.93  & & 58.74 & 52.81\\
3 & \multicolumn{3}{c}{Every $5$ phases} && 63.46 & 64.91  & & 58.67 & 52.42\\
\bottomrule
\end{tabular}}
}
\end{center}
\vspace{-0.5cm}
\caption{Ablation results on CIFAR-100, including the average accuracy across all phases. The baseline is LUCIR~\cite{hou2019lucir} and No.4 shows our best result.} 
\vspace{-0.2cm}
\label{table_ablation_supp}
\end{table}

%% file: tables/time.tex
\setlength{\tabcolsep}{0.4mm}{\begin{table}[ht!]
  \small
  \centering
  \vspace{0.1cm}
  {\renewcommand\arraystretch{1.1}
  \begin{tabular}{lcccccccccccc}
  \toprule
    \multirow{2.5}{*}{Method}&\multirow{2.5}{*}{Baseline}&\multirow{1.5}{*}{Bilevel}&\multirow{1.5}{*}{Offline}&  \multicolumn{3}{c}{\textsl{w/} Ours}\\
    \cmidrule{5-7}
    && HO&RL&1 phase& 2 phases&5 phases\\
    \midrule
    Time (hours)&1.6&3.0&650&5.2&3.8&2.3\\
  \bottomrule

\end{tabular}}
		\centering
		\vspace{-0.1cm}
	\caption{Average training time in TFH on CIFAR-100 $5$-phase using an NVIDIA v100 GPU. The baseline is LUCIR~\cite{hou2019lucir}. The offline RL method is based on the pseudo task RL introduced by \cite{Liu2021RMM}. The Bilevel HO method is based on \cite{franceschi2018bilevel}. We show the training time of our method with different policy updating frequencies: every 1/2/5 phases.}
	\label{table_ablation_time}	
	\vspace{-0.2cm}
\end{table}}